\documentclass{article} % For LaTeX2e
\usepackage{iclr2019_conference,times}
\usepackage{mathrsfs,amsmath}
\usepackage{amsthm}
\usepackage{paralist}
\usepackage{amsmath,amsfonts,bm}

\def\eqref#1{equation~\ref{#1}}
\def\1{\bm{1}}

\DeclareMathAlphabet{\mathsfit}{\encodingdefault}{\sfdefault}{m}{sl}
\SetMathAlphabet{\mathsfit}{bold}{\encodingdefault}{\sfdefault}{bx}{n}

\usepackage{hyperref}
\usepackage{url}
\usepackage{svg}
\usepackage{booktabs}       % professional-quality tables
\usepackage{hyperref}

\newtheorem{theorem}{Theorem}[section]

\newtheorem{definition}{Definition}[section]

\usepackage{xcolor}

\newcommand{\ie}{\textit{i}.\textit{e}., }
\newcommand{\eg}{\textit{e}.\textit{g}., }

\title{edGNN: A Simple and Powerful GNN for Directed Labeled Graphs}

\author{Guillaume Jaume\thanks{equal contribution} \\
IBM Research, Zurich \\
EPFL, Lausanne \\
\texttt{gja@zurich.ibm.com} \\
\And
An-phi Nguyen\footnotemark[1] \\
IBM Research, Zurich \\
ETH, Zurich \\
\texttt{uye@zurich.ibm.com} \\
\And 
Mar\'{i}a Rodr\'{i}guez Mart\'{i}nez \\
IBM Research, Zurich \\
\texttt{mrm@zurich.ibm.com}
\And 
Jean-Philippe Thiran \\
EPFL, Lausanne \\
\texttt{jean-philippe.thiran@epfl.ch} \\
\And
Maria Gabrani \\
IBM Research, Zurich \\
\texttt{mga@zurich.ibm.com} \\
}

\iclrfinalcopy % Uncomment for camera-ready version, but NOT for submission.
\begin{document}

\maketitle

\begin{abstract}

% Graphs are an ubiquitous way to represent structured data.
The ability of a graph neural network (GNN) to leverage both the graph topology and graph labels is fundamental to building discriminative node and graph embeddings. Building on previous work, we theoretically show that edGNN, our model for directed labeled graphs, is as powerful as the Weisfeiler--Lehman algorithm for graph isomorphism. Our experiments support our theoretical findings, confirming that graph neural networks can be  used effectively for inference problems on directed graphs with both node and edge labels. Code available at \url{https://github.com/guillaumejaume/edGNN}.

\end{abstract}

\section{Introduction}

%Graphs are an ubiquitous way to represent structured data, such as ontologies, social networks, protein interaction networks just to name a few. Several are the challenging tasks that can be performed on this kind of data. For example, an important application in cancer research is the prediction of the carcinogenicity of a chemical compound solely based on its structure (\cite{Helma2001}). In social science, it is interesting to understand if the social circle of an user may influence its political standing (\cite{Stanford}).  It is therefore not surprising that part of the research community have focused its efforts to develop models for graph data, from spectral theory-based (\cite{ilprints422}) to kernel-based methods (\cite{Shervashidze2009}).

In recent years, much work has been devoted to extending deep-learning models to graphs,~\eg~\cite{Scarselli2009, Bruna, Li2015b, Defferrard, Kipf2017, Hamilton2017, Hamilton2017a, Velickovic2017a, Ying2018a}. \cite{Gilmer2017a} formulated numerous such models within their proposed Message Passing Neural Network (MPNN) framework. In this model, the state of a vertex is represented by a feature vector. The state is then  updated iteratively via a two-step strategy: For each vertex \begin{inparaenum}[(i)]
\item in the aggregation step, the feature vectors of the neighboring vertices are combined into a single vector via a differentiable operator, \eg a sum. Then, 
\item in the update step, a new state is computed by applying another differentiable operator, \eg a one-layer perceptron, to the current state of the vertex and to the aggregate vector from the previous step.
\end{inparaenum} 

In two recent works, \cite{Xu2018} and \cite{Morris2018}, independently proved that certain formulations of MPNNs are \emph{as powerful as} the Weisfeiler--Lehman (WL) algorithm for graph isomorphism~(\cite{Weisfeiler1968}). 
%This result is similar, in spirit, to the Universal Approximation Theorem for neural networks (\cite{Cybenkot1989}). 
In practice, this means that there exist MPNNs able to learn \emph{unique} representations for 
(\emph{almost})
%(\emph{almost} \footnote{The WL algorithm, although widely used, is not able to distinguish among all non-isomorphic graphs.})
\emph{all}  undirected node-labeled graphs, which is desirable for such tasks as node or graph classification. 
% @TODO: we don't obvisouly have to keep this
Note that previous work has drawn a parallel between the WL test and the MPNN, \eg ~\cite{Hamilton2017, Jin2017, Lei2017}.

In this paper, we extend the above-mentioned results to \emph{directed} graphs with labels for both nodes and \emph{edges}. In particular, by extending the theoretical framework provided by \cite{Morris2018}, we show that there exist MPNNs as powerful as the (one-dimensional) WL algorithm for directed labeled graphs. Although this problem has already been addressed,~\eg~\cite{Li2015b, Niepert2016, Simonovsky2017, Beck2018, Schlichtkrull2018}, we present a theoretically-grounded GNN formulation for directed labeled graphs. We experimentally corroborate our theoretical results by comparing our model, edGNN, against state-of-the-art models for node and graph classification.
\section{Theoretical framework}
\subsection{Notation and setup}

As our work is an extension of~\cite{Morris2018}, we maintain a notation consistent to theirs in order to more easily reference their results. 

A graph $G$ is a pair $(V,E)$, where $V$ is the set of vertices, $E$ is the set of edges and the directed edge $(u,v) \in E$ for $u, v \in V$ is an edge starting in $u$ and ending in $v$. We will denote the vertex and the edge sets of $G$ as $V_G$ and $E_G$, respectively.

We are interested in graphs with both node and edge labels. We therefore assume that, given a graph $G$, there exists a vertex-labeling function $l_V: V_G \rightarrow \mathcal{X}$ and an edge-labeling function $l_E: E_G \rightarrow \mathcal{Z}$ that assign to each vertex and edge of $G$ a label from \emph{countable} sets $\mathcal{X}$ and $\mathcal{Z}$. For the rest of this paper, we will refer to graphs with node and edge labels simply as \emph{labeled graphs}.
%Note that for ease of notation and without loss of generality the label sets can be taken to be equal, i.e. $\mathcal{X} = \mathcal{Z}$. 

For each vertex $v \in V$, we can define the neighborhood $\mathcal{N}(v) := \{u \in V \; | \; (v,u) \in E \; \vee \; (u,v) \in E\}$. As we are dealing with directed graphs, we distinguish between incoming neighbors $\mathcal{N}^I(v) := \{u \in V \; | \; (u,v) \in E\}$ and outgoing neighbors $\mathcal{N}^O(v) := \{u \in V \; | \; (v,u) \in E\}$. Naturally, $\mathcal{N}(v) = \mathcal{N}^I(v) \cup \mathcal{N}^O(v)$. The cardinalities $|\mathcal{N}(v)|, \; |\mathcal{N}^I(v)| \; \mathrm{and} \; |\mathcal{N}^O(v)|$ of the neighborhoods are referred to as the \emph{degree}, the \emph{in-degree} and the \emph{out-degree} of vertex $v$, respectively.

\begin{definition}
Two labeled directed graphs $G$ and $H$ are \emph{isomorphic} if there exists a bijection $f: V_G \rightarrow V_H$ such that $(u, v) \in E_G$ if and only if $(f(u), f(v)) \in E_H$ with $l_{V_G}(v) = l_{V_H}(f(v))$ and $l_{E_G}(u, v) = l_{E_H}(f(u), f(v))$. 
\end{definition} 

\subsubsection{The Weisfeiler--Lehman algorithm}\label{sec:wl_test}

The Weisfeiler--Lehman (WL) test~(\cite{Weisfeiler1968}) is an algorithm to distinguish whether two graphs are non-isomorphic. We present the test in its one-dimensional variant, also known as the \emph{naive vertex refinement}. We will start by presenting the WL test on node-labeled graphs, and later discuss its extension to directed labeled graphs. 

At initialization, the vertices are labeled consistently with the vertex-labeling function $l_V$. We call this the \emph{initial coloring} of the graph and we denote it as $c_l^{(0)}(v) := l_V(v), \; \forall v \in V$. The algorithm then proceeds in a recursive fashion. At iteration $t$, new labels are computed for each vertex from the current labels of the vertex itself and its neighbors,~\ie
\begin{align}\label{eq:wl_iteration}
    c_l^{(t)}(v) = g\Big(c_l^{(t-1)}(v), \big\{\big\{ c_l^{(t-1)}(u): u \in \mathcal{N}(v) \big\}\big\}\Big),
\end{align}
where $g$ is an injective hashing function, and $\big\{\big\{\cdot\big\}\big\}$ denotes a \emph{multiset},~\ie a generalization of a set that allows elements to be repeated.
Each iteration is performed in parallel for the two graphs to be tested, $G$ and $H$. If at some iteration $t$, the number of vertices assigned to a label $l \in \mathcal{X}$ differs for the two graphs, then the algorithm stops, concluding that the two graphs are not isomorphic. Otherwise, the algorithm will stop whenever a \emph{stable coloring} is achieved,~\ie whenever $c_l^{(t)}(v_G) = c_l^{(t)}(v_H)$ for all $t \geq T$ and for any pair $(v_G, v_H)$ with $v_G \in V_G$, $v_H \in V_H$, and $c_l^{(T)}(v_G) = c_l^{(T)}(v_H)$.
This is guaranteed to happen at most after $T=\mathrm{max}\{|V_G|,|V_H|\}$ iterations.
In this case, $G$ and $H$ are considered isomorphic. 

% @TODO eventually cut this paragraph 
% @TODO if we keep, address citation issue 
Despite a few corner cases, the WL test is able to distinguish a wide class of graphs~(\cite{Cai1992}). Moreover, the most efficient implementation of the algorithm has a runtime complexity which is quasi-linear in the number of vertices~(\cite{Grohe2017a%,CARDON198285,Paige:1987:TPR:37185.37186
}).
%For these reasons, the WL test is still commonly used as an isomorphism test.
% This sentence is redundant with the 1st one of the paragraph. 

The extension of the WL test to a directed graph with edge labels is straightforward~(\cite{Grohe2017a, Orsini2015}). During the recursive step, for each vertex $v$, we need to include  the in-degrees and  out-degrees of $v$ separately in the hashing function \emph{with respect to each} edge label.

Let us denote an edge label as $e \in \mathcal{Z}$. For each vertex $v$, we then define $n^I_v(e) := |\{u \in \mathcal{N}^I(v) \; | \; l_E(u,v) = e\}|$ as the number of edges incoming to $v$ with label $e$. Similarly, $n^O_v(e)$ is defined for outgoing edges. Then,  Eq.~(\ref{eq:wl_iteration}) can be adapted for labeled directed graphs in the following way:
\begin{equation}\label{eq:wl_iteration_dir_edgelabel}
\begin{split}
    c_l^{(t)}(v) = g\Big(c_l^{(t-1)}(v), &\big\{\big\{ c_l^{(t-1)}(u): u \in \mathcal{N}(v) \big\} \big\}, \\ &\big\{ \big(n^I_v(e), e\big): \exists (u,v) \in E\;\mathrm{with}\;l_E(u,v) = e \big\},\\ &\big\{ \big(n^O_v(e), e\big): \exists (v,u) \in E\;\mathrm{with}\;l_E(v,u) = e \big\}\Big).
\end{split}
\end{equation}

\subsection{Graph neural networks}

% @TODO rm this part? As noted by \cite{Gilmer2017a}, modern g
Graph neural networks architectures implement a neighborhood aggregation strategy. Similar to~\cite{Morris2018}, our work focuses on the GNN model presented by \cite{Hamilton2017}, which implements this strategy with the node update function 
\begin{align}\label{eq:1gnn}
    f^{(t)}(v) = \sigma\Big(f^{(t-1)}(v)W_1^{(t)} + \sum_{u \in \mathcal{N}(v)}f^{(t-1)}(u)W_2^{(t)}\Big),
\end{align}

where $f^{(t)}(v) \in \mathbb{R}^{1 \times d^{(t)}}$ is the $d^{(t)}$-dimensional node representation, or node embedding, at time step $t$ of the vertex $v$, and $W_1^{(t)}, W_2^{(t)} \in \mathbb{R}^{d^{(t-1)}\times d^{(t)}}$ are weight matrices. The initial representation $f^{(0)}(v)$ is \emph{consistent} with the vertex-labeling function $l_V$, \ie~$f^{(0)}(v) = f^{(0)}(u)$ if and only if $l_V(v) = l_V(u)$ for all $v,u \in V$. \cite{Morris2018} showed that there exists a sequence $(W_1^{(t)},W_2^{(t)})$ such that the GNN model in Eq.~(\ref{eq:1gnn}) is as powerful as the WL test.
% @TODO talk about the readout already here?

\subsubsection{Extension to directed labeled graphs}

The extension of Eq.~(\ref{eq:1gnn}) to directed labeled graphs follows the WL test extension. We simply need to augment the equation with embeddings for the labeled edges with incoming and outgoing edges considered separately,~\ie
\begin{equation}\label{eq:1gnn_edges}
\begin{split}
    f^{(t)}(v) = \sigma\Big(&f^{(t-1)}(v)W_1^{(t)} + \sum_{u \in \mathcal{N}(v)}f^{(t-1)}(u)W_2^{(t)} + \\ &+ \sum_{(u,v) \in E}f_E\big(u,v, l_E(u,v)\big)W_3^{(t)} + \sum_{(v,u) \in E}f_E\big(v,u, l_E(v,u)\big)W_4^{(t)}\Big),
\end{split}
\end{equation}

where $f_E\big(v,u,l_E(v,u)\big) \in \mathbb{R}^{1 \times d_E}$ is the $d_E$-dimensional embedding of the edge $(v,u)$ with label $l_E(v,u)$. The embeddings $f_E$ should be defined such that $\sum_{(u,v) \in E}f_E\big(u,v, l_E(u,v)\big) = \sum_{(u,v') \in E}f_E\big(u,v', l_E(u,v')\big)$ if and only if $\big\{ \big(n^I_v(e), e\big): \exists (u,v) \in E\;\mathrm{with}\;l_E(u,v) = e \big\} = \big\{ \big(n^I_{v'}(e), e\big): \exists (u,v') \in E\;\mathrm{with}\;l_E(u,v') = e \big\}$. The same should hold for outgoing edges. For practical applications, this can be achieved by using one-hot encodings of the edge labels.\\
We can now directly extend the theorems presented in~\cite{Morris2018}.
\begin{theorem}[Theorem 1 in \cite{Morris2018}]\label{thm:at_most} Let G be a directed labeled graph. Then for all $t\geq0$ and for all choices of initial colorings $f^{(0)}$ consistent with $l_V$ and of edge embeddings $f_E$ consistent with $l_E$, and weights $W_1^{(t)}, W_2^{(t)},W_3^{(t)},W_4^{(t)}$
\begin{align}\label{eq:refines}
    c_l^{(t)}(v) = c_l^{(t)}(u) \Rightarrow f^{(t)}(v) = f^{(t)}(u) \;\; \forall u,v \in V
\end{align}
with $c_l^{(t)}$ and $f^{(t)}$ defined in Eqs.~(\ref{eq:wl_iteration_dir_edgelabel}) and (\ref{eq:1gnn_edges}),  respectively. 
%where Eq.~(\ref{eq:refines}) means that $c_l^{(t)}(v) = c_l^{(t)}(u) \Rightarrow f^{(t)}(v) = f^{(t)}(u) \;\; \forall u,v \in V$.
\end{theorem}
\cite{Morris2018} prove this theorem by induction. The proof is essentially the same for our extended case. In fact, as neither the labels $l_E$ nor the embeddings $f_E$ change over the iterations, there is no need to include them in the induction step.
\begin{theorem}[Theorem 2 in \cite{Morris2018}]\label{thm:exists} Let G be a directed labeled graph with finite vertex degree. Then there exists a sequence $(W_1^{(t)}, W_2^{(t)},W_3^{(t)},W_4^{(t)})$ with $t\geq0$ such that
\begin{align}\label{eq:equiv}
     c_l^{(t)}(v) = c_l^{(t)}(u)  \Leftrightarrow f^{(t)}(v) = f^{(t)}(u) \;\; \forall u,v \in V
\end{align}
%where Eq.~(\ref{eq:equiv}) means that $c_l^{(t)} \sqsubseteq f^{(t)}$ and $f^{(t)} \sqsubseteq c_l^{(t)}$.
\end{theorem}

The proof is provided in the supplemental material. Note that we specifically require the graph to have finite vertex degree. However, this is not a strong assumption in real-world applications.

\subsubsection{Graph classification}

For graph classification tasks, we need a representation $f_G$ of the graph. We build it from the node representations $f^{(t)}(v)$ following the formulation of \cite{Xu2018}:  % @TODO justify the use of a sum ? (over mean or max)
\begin{align}
    f_G = \mathrm{Concat}\Big( \Big\{\sum_{v \in V_G} f^{(t)}(v) \;\Big|\;t = 0, \dots , T\Big\}\Big).
\end{align}
Note that, although  $T$ should theoretically be at least $|V_G|$ (Section~\ref{sec:wl_test}), only a few layers (\ie iterations) are used in practice to update the node representations. Finally, a linear classifier is  applied to the graph representation to perform the classification.

\section{Experiments}

\subsection{Datasets and baselines}

% We benchmark our algorithm edGNN on node and graph classification tasks.
% For node classification tasks, we compare our model against R-GCN~(\cite{Schlichtkrull2018}), RDF2Vec~(\cite{Ristoski2016a}) and WL~(\cite{Shervashidze2011, DeVries2015}) on the AIFB and MUTAG dataset~(\cite{Ristoski2016b}). For graph classification tasks, we benchmark our model against the Subgraph Matching Kernel (CSM)~(\cite{Kriege2012}), Weisfeiler--Lehman Shortest Path Kernel~(\cite{Shervashidze2011}) and R-GCN~(\cite{Schlichtkrull2018}). As R-GCN only defines how to build node embeddings, we reuse the formulation of \cite{Xu2018} to build a graph-level representation.
% Dataset statistics as well as training details are reported in the Appendix.

We benchmark our algorithm edGNN on graph and node classification tasks.
For graph classification tasks, we benchmark our model against the Subgraph Matching Kernel (CSM)~(\cite{Kriege2012}), Weisfeiler--Lehman Shortest Path Kernel~(\cite{Shervashidze2011}) and R-GCN~(\cite{Schlichtkrull2018})
%on the MUTAG~(\cite{Debnath1991,Kriege2012}) and PTC~(\cite{Helma2001}) datasets
. As R-GCN only defines how to build node embeddings, we reuse the formulation of \cite{Xu2018} to build a graph-level representation. For node classification tasks, we compare our model against R-GCN~(\cite{Schlichtkrull2018}), RDF2Vec~(\cite{Ristoski2016a}) and WL~(\cite{Shervashidze2011, DeVries2015}) on the AIFB and MUTAG dataset~(\cite{Ristoski2016b}). 
Dataset statistics as well as training details are reported in the Appendix.

\subsection{Results and discussion}

In Table~\ref{table:result}, we report results for the graph (left-hand side) and node classification (right-hand side) tasks. Our provably powerful model, edGNN, reaches comparable performance with the state-of-the-art. We observe that the kernel-based and gradient-based methods (R-GCN and edGNN) perform similarly without being able to clearly identify better models. We conjecture that the relatively small size of the datasets (\eg only $188$ graphs in the MUTAG dataset) does not allow to fully explore the potential of the most expressive models.
%% Change this paragraph. 
% In the case of graph-kernel-based methods, we conjecture that the lower performances are due to the use of less powerful models, which could lead to their underfitting. Both CSM and WLSP focus on specific subgraph features (\eg shortest paths), which may not be sufficient to fully characterize a graph in the datasets we analyzed.
% However, as we have no information about the training accuracies for these models, we cannot confirm this hypothesis.
% In the case of R-GCNs, as there is no theoretical result proving that they can or \emph{cannot} be equivalent to the WL test, the consistently low performance might merely be because the R-GCN formulation is not appropriate for the particular learning setting of our experiments (\eg relatively small datasets).
% node classification
% For node classification (right-hand table in Table~\ref{table:result}), edGNN also achieves comparable performance with the state-of-the-art without outperforming it.
This does not contradict our theoretical findings. In fact, the power of a learnable model does not guarantee its generalization nor that the best model can be learned. However, it is true that, \emph{in the best-case scenario}, a more powerful model should perform better than a less powerful one, as shown by the results regarding the best-learned edGNN model (max).
% future direction
An interesting future research direction would be the study of all the proposed models fitting the MPNN framework in order to understand whether they can be as powerful as the WL test or whether, instead, they introduce a particular bias. 

\begin{table}[h]
\caption{
Graph (left) and node (right) classification results in accuracy averaged over ten runs. For the proposed edGNN approach, we report both the average (avg) and the maximum accuracy (max). Results are expressed as percentages.
For graph classification, following prior art, we performed 10-fold cross validation. For node classification, we used the split provided by~\cite{Ristoski2016a}.
}
\label{table:result}
\centering
\scalebox{0.8}{
\begin{tabular}{lccccc}
  \toprule
    Model & MUTAG & PTC FM & PTC FR & PTC MM & PTC MR \\
    \midrule
    CSM         & $85.4$          & $\mathbf{63.8}$ & $65.5$          & $63.3$          & $58.1$ \\
    WLSP        & $85.4$          & $60.4$          & ${65.7}$ & $\mathbf{66.6}$ & $\mathbf{59.7}$ \\
    R-GCN       & $81.5$          & $60.7$          & $\mathbf{65.8}$          & $64.7$          & $58.2$ \\
    \midrule
    edGNN (avg) & $\mathbf{86.9}$ & $59.8$          & ${65.7}$ & $64.4$          & $56.3$ \\
    edGNN (max) & $88.8$          & $62.2$          & $68.0$          & $66.1$          & $59.4$ \\
    \bottomrule
\end{tabular}
\quad
\begin{tabular}{lccc}
  \toprule
    Model & AIFB & MUTAG \\
    \midrule
    WL &   $80.5$  & $\mathbf{80.9}$  \\
    RDF2Vec & $88.9$ & $67.2$  \\
    R-GCN & $\mathbf{95.8}$ & $73.2$ \\
    \midrule
    edGNN (avg) & $91.1$ & $80.0$  \\
    edGNN (max) & $97.2$ & $85.3$ \\
    %\midrule
    %edGNN (emb) &  $91.1$ & $77.2$ \\
    %edGNN (reg) &  $89.4$ & $80.4$ \\
    \bottomrule
  \end{tabular}
}
\end{table}

%\begin{itemize}
%    \item Table 1 left: edGNN max have the best results supporting that we are more powerful, which does not mean though that we are better on average, i.e. power does not mean generalization (or good performance). 
%    \item Table 1 right (edGNN vs RGCN): edGNN has consistently better results than other methods, one more experimental evidence for power. according to authors, RGCN is designed to specifically deal with highly multi-relational blah blah (cite) (approximation). WLSP and CSM
%    \item Table 1 right (edGNN vs WLSP): WLSP only computes shortest paths as features. this might not enough for these graph classification tasks.
%\end{itemize}

%\section{Conclusion}

%We introduce edGNN, a graph neural network model which leverages edge labels. We theoretically proved that this formulation, whilst simple, can be as powerful as the well-established Weisfeiler-Lehman algorithm for graph isomorphism. That is, our model has the same ability of the WL test in distiguishing non-isomorphic directed graphs with labeled nodes and edges. We experimentally demonstrated that edGNN is indeed able to achieve state-of-the-art results on graph and node classification benchmark datasets. 

% papers to cite that are benchmarking on the PTC graph classification dataset are:
% %
% \begin{enumerate}
%     \item A degeneracy framework for graph similarity~\cite{Nikolentzos2018}
%     \item Fast depth-based subgraph kernels for unattributed graphs~\cite{Bai2016}
%     \item An End-to-End Deep Learning Architecture for Graph Classification~\cite{Zhangb}
% \end{enumerate}

\clearpage
\bibliography{GINe}
\bibliographystyle{iclr2019_conference}

\clearpage
\section{Appendix}

\subsection{Proof of Theorem~\ref{thm:exists}} 

For a given vertex $v$, we define $E^I_v := \{ l_E(u,v):  \exists (u,v) \in E\}$ as the set of labels of edges incoming into the vertex $v$. We then define $L_v^I := \mathrm{Concat}\Big(\big\{ \big(n^I_v(e), e\big): \forall e \in E^I_v \big\}\Big)$. That is, for each vertex $v$, we create a label by concatenating all the labels of the incoming edges together with their multiplicities $n^I_v(e)$. Similarly, we define $E^O_v$ and $L_v^O$ for the outgoing edges. Note that the pairs $\big(n^I_v(e), e\big)$ and $\big(n^O_v(e), e\big)$ take values in $\mathcal{L} := \mathbb{N} \times \mathcal{Z}$. Therefore, $L_v^I$ (or $L_v^O$, respectively) can take values in $\mathcal{L}^{|E^I_v|}$ (or $\mathcal{L}^{|E^O_v|}$, respectively), \ie~the Cartesian product of $|E^I_v|$ (or $|E^O_v|$, respectively) copies of the set $\mathcal{L}$.

For all vertices $v$, we can construct a function $h_v: \mathcal{X} \times \mathcal{L}^{|E^I_v|} \times \mathcal{L}^{|E^O_v|} \rightarrow \mathcal{Y}_v$ that \emph{bijectively} maps a tuple $\Big(c_l^{(t-1)}(v),L^I_v,L^O_v\Big)$ to a label $y \in \mathcal{Y}_v$. 

Note that, as we are considering graphs with finite vertex degree, $|\mathcal{N}^I(v)|$ and $|\mathcal{N}^O(v)|$ (and, consequently, $|E^I_v|$ and $|E^O_v|$) are finite. Therefore, $\mathcal{X} \times \mathcal{L}^{|E^I_v|} \times \mathcal{L}^{|E^O_v|}$ is a \emph{countable} set because the finite Cartesian product of countable sets is itself countable. Thus, as we built the function $h_v$ to be bijective, the sets $\mathcal{Y}_v$, and their countable union $\mathcal{Y} := \bigcup_{v \in V} \mathcal{Y}_v$, are also countable (for results on countable sets, refer for example to~\cite{Patterson1967}).

We can then construct an injective hash function $g'$ such that 
\begin{equation}\label{eq:reformulate_wl}
\begin{split}
    g'\Big(y, \big\{ c_l^{(t-1)}(u): u \in \mathcal{N}(v) \big\}\Big)
    = g\Big(c_l^{(t-1)}(v), &\big\{ c_l^{(t-1)}(u): u \in \mathcal{N}(v) \big\},L^I_v,L^O_v\Big).
\end{split}
\end{equation}
where the right-hand side is the relabeling function defined in Eq.~(\ref{eq:wl_iteration_dir_edgelabel}). 

These constructions highlight the fact that an iteration of the WL algorithm on a directed labeled graph is the same as performing an iteration of the WL algorithm on a undirected node-only-labeled graph, where node labels take values in an appropriately augmented label set $\mathcal{Y}$.

The same equivalence can be highlighted between the GNN update functions in Eqs.~(\ref{eq:1gnn}) and (\ref{eq:1gnn_edges}). In fact, Eq.~(\ref{eq:1gnn_edges}) can be rewritten as 
\begin{align}\label{eq:reformulate_1gnn}
    f^{(t)}(v) = \sigma\Big(f_\mathcal{Y}^{(t-1)}(v)W_{1,3,4}^{(t)} + \sum_{u \in \mathcal{N}(v)}f^{(t-1)}(u)W_2^{(t)}\Big),
\end{align}

where $f_\mathcal{Y}^{(t)}(v) \in \mathbb{R}^{1 \times (d^{(t)} + 2d_E)}$ is the embedding resulting from the (horizontal) concatenation of $f^{(t)}(v)$, $\sum_{(u,v) \in E}f_E\big(u,v, l_E(u,v)\big)$, and $\sum_{(v,u) \in E}f_E\big(v,u, l_E(v,u)\big)$, whereas $W_{1,3,4}^{(t)}$ is the (vertical) concatenation of $W_{1}^{(t)}$, $W_{3}^{(t)}$, and $W_{4}^{(t)}$.

The reformulations presented in Eqs.~(\ref{eq:reformulate_wl}) and  (\ref{eq:reformulate_1gnn}) allow us to treat our problem as one of undirected graphs with labels only for nodes. We can therefore prove this theorem by directly using the proof of Theorem 2 in \cite{Morris2018}. $\qed$

\subsection{Details of training and more results}

\subsection{Initialization}

We initialize the node and edge features with a one-hot encoding of their input label. For node classification, we use the in-degree as input label of the nodes. To model the outgoing edges of the node classification graphs, we create new artificial relations by reversing each directed edge. We also report results without reversing the edges (reg). and with learned embeddings (emb) instead of a one-hot encoding.

\subsubsection{Graph classification}

The tasks  predict the mutagenicity~(\cite{Debnath1991,Kriege2012}) (MUTAG) and the toxicity~(\cite{Helma2001}) of chemical compounds (PTC).

We ran our graph classification experiments with a batch size of $8$ and a learning rate of $10^{-4}$ with a weight decay of $5 \times 10^{-4}$. We then performed a parameter search over the number of layers and node embedding size. The best performance was reached by using two layers and $64$ hidden units. We used a ReLu activation at each layer. The system was trained for at most $40$ epochs with early stopping with respect to the validation set cross-entropy loss. 

R-GCN for graph classification was also trained with a batch size of $8$, a learning rate of $10^{-4}$ with a weight decay of $5 \times 10^{-4}$. We used three layers and $64$ hidden units with learned initial embeddings for the nodes (instead of a one-hot encoding for edGNN).
We used a basis decomposition with the number of basis set to the number of relations. Results for CSM~(\cite{Kriege2012}) and WLSP~(\cite{Shervashidze2011}) are based on the re-implementation of \cite{Kriege2012}. All our experiments were performed with 10-fold cross validation as in~\cite{Kriege2012}. 

Dataset statistics for the graph classification tasks are shown in Table~\ref{tab:graph_class_stat}, whereas Table~\ref{tab:graph_class} shows the accuracy results together with the standard deviation. 

\begin{table}[h]
  \caption{Graph classification dataset statistics with the number of graphs (Graphs), the number of classes (Classes), the average number of nodes per graph (Avg nodes), the average number of edges per graph (Avg edges), the number of node labels (Node labels) and the number of edge labels (Edge labels).}
  \label{tab:graph_class_stat}
  \centering
  \begin{tabular}{lcccccc}
  \toprule
    Dataset & Graphs & Classes & Avg nodes & Avg edges & Node labels & Edge labels   \\
    \midrule
    MUTAG  &   188  & 2 & 17.9 & 19.8  & 6 & 3   \\
    PTC FM &   349  & 2 & 14.1 & 14.5 & 18 & 4  \\
    PTC FR &   351  & 2 & 14.6 & 15.0 & 19 & 4  \\
    PTC MM &   336  & 2 & 14.0 & 14.3 & 20 & 4  \\
    PTC MR &   344  & 2 & 14.3 & 14.7 & 18 & 4  \\
    \bottomrule
  \end{tabular}
\end{table}

\begin{table}[h]
  \caption{Graph classification results in accuracy obtained with 10-fold cross validation. Results are expressed as percentages. edGNN is compared with the Subgraph Matching Kernel (CSM)~(\cite{Kriege2012}), Weisfeiler--Lehman Shortest Path Kernel~(\cite{Shervashidze2011}) and R-GCN~(\cite{Schlichtkrull2018}).}
  \label{tab:graph_class}
  \centering
  \begin{tabular}{lccccc}
  \toprule
    Model & MUTAG & PTC FM & PTC FR & PTC MM & PTC MR \\
    \midrule
    CSM & ${85.4 \pm 1.2}$  & $\mathbf{63.8 \pm 1.0}$ & $65.5 \pm 1.4$  & $63.3 \pm 1.7$ & $58.1 \pm 1.6$ \\
    WLSP & ${85.4 \pm 1.2}$ & $60.4 \pm 1.32$ & ${65.7 \pm 1.3}$ & $\mathbf{66.6 \pm 1.1}$ & $\mathbf{59.7 \pm 1.6}$ \\
    R-GCN & $81.5 \pm 2.1$  & $60.7 \pm 1.7$  & $\mathbf{65.8 \pm 0.6}$ & $64.7 \pm 1.7$  & $58.2 \pm 1.7$ \\
    \midrule
    edGNN (avg) & $\mathbf{86.9 \pm 1.0}$ & $59.8 \pm 1.5$          & ${65.7 \pm 1.3}$ & $64.4 \pm 0.8$          & $56.3 \pm 1.9$ \\
    edGNN (max) & $88.8$          & $62.2$          & $68.0$          & $66.1$          & $59.4$ \\
    
    \bottomrule
  \end{tabular}
\end{table}

\subsubsection{Node classification}

The node classification experiments were run with a learning rate of $5 \times 10^{-3}$  without weight decay. We used dropout $0.5$ on each layer with a ReLu activation. The best performance was achieved by using $2$ layers and $64$ hidden units. The maximum number of epochs was set to $400$ with early stopping with respect to the validation set cross-entropy loss. Results with R-GCN~(\cite{Schlichtkrull2018}), RDF2Vec~(\cite{Ristoski2016a}) and WL~(\cite{Shervashidze2011}) are based on the re-implementation of \cite{Schlichtkrull2018}. 

\begin{table}[h]
  \caption{Node classification dataset statistics with the number of classes (Classes), the total number of nodes (Nodes), the total number of edges (Edges) and the number of edge labels (Edge labels).}
  \label{tab:node_class_stat}
  \centering
  \begin{tabular}{lccccc}
  \toprule
    Dataset & Classes & Nodes & Edges & Edge labels   \\
    \midrule
    AIFB  &   4 & 8,285  & 29,043  & 45  \\
    MUTAG &   2 & 23,644 & 74,227  & 23  \\
    \bottomrule
  \end{tabular}
\end{table}

\begin{table}[h]
  \caption{Node classification results in accuracy averaged over ten runs. Results are expressed as percentages. edGNN is compared with WL~(\cite{DeVries2015}), RDF2Vec~(\cite{Ristoski2016a}) and R-GCN~(\cite{Schlichtkrull2018}).}
  \label{tab:node_class}
  \centering
  \begin{tabular}{lccc}
  \toprule
    Model & AIFB & MUTAG \\
    \midrule
    WL          & $80.5 \pm 0.0$          & $\mathbf{80.9 \pm 0.0}$  \\
    RDF2Vec     & $88.9 \pm 0.0$          & $67.2\pm 1.2$            \\
    R-GCN       & $\mathbf{95.8 \pm 0.6}$ & $73.2 \pm 0.5$           \\
    edGNN (avg) & $91.1 \pm 2.4$          & $80.0 \pm 3.2$           \\
    edGNN (max) & $\mathbf{97.2}$         & $\mathbf{85.3}$          \\
    \midrule
    edGNN (emb) & $91.1 \pm 1.7$          & $77.2 \pm 2.6$           \\
    edGNN (reg) & $89.4 \pm 1.7$          & $80.4 \pm 3.4$           \\
    \bottomrule
  \end{tabular}
\end{table}

\end{document}